\documentclass[10pt, a4paper]{article}
\usepackage{lrec2006}
\usepackage{graphicx}
\usepackage[bf,small]{caption}
\usepackage{amsmath}
\usepackage{url}
\newcommand{\argmax}{\operatornamewithlimits{argmax}}

\title{A Light Sliding-Window Part-of-Speech Tagger for the Apertium Free/Open-Source Machine Translation Platform}

\name{Gang Chen, Mikel L. Forcada}

\address{ Key Laboratory of Computational Linguistics (Peking University), Ministry of Education, China\\
          Departament de Llenguatges i Sistemes Inform{\`a}tics, Universitat d'Alacant, Spain \\
          pkuchengang@gmail.com, mlf@dlsi.ua.es\\}

\abstract{This paper describes a free/open-source implementation of the light sliding-window (LSW) part-of-speech tagger for the Apertium free/open-source machine translation platform. Firstly, the mechanism and training process of the tagger are reviewed, and a new method for incorporating linguistic rules is proposed. Secondly, experiments are conducted to compare the performances of the tagger under different window settings, with or without Apertium-style ``forbid'' rules, with or without Constraint Grammar, and also with respect to the traditional HMM tagger in Apertium. \\
}

\begin{document}

\maketitleabstract


\section{Introduction}
Apertium\footnote{The Apertium machine translation engine, linguistic data for various language pairs, and documentation can be downloaded from \url{http://www.apertium.org}.} is a shallow-transfer rule-based free/open-source machine translation platform. This paper reports a free/open-source implementation of the light sliding window (LSW) PoS tagger \cite{Enrique-05}, and compares its performance with that of the original first-order HMM tagger in Apertium \cite{Francis-10,Sheikh-09,Doug-92}. Section 2 reviews the mechanism of the LSW tagger and proposes a method to improve its tagging accuracy by incorporating linguistic rules, Section 3 shows the experimental results and discusses them, and finally, in Section 4, the paper ends with some conclusions and future plans.


\section{Methods}
The main difference between the LSW and HMM PoS taggers is that the LSW PoS tagger makes local decisions about the PoS tag of each word which are based on the ambiguity class (set of PoS tags) of words in a fixed-length context around the problem word, while HMM makes this decision by efficiently considering all possible disambiguations of all words in the sentence, by using a probabilistic model based on a multiplicative chain of transition and emission probabilities. In terms of model complexity, LSW is simpler than HMM, while, on the other hand, the number of parameters of LSW could be larger than that of HMM, which may have a crucial influence on the tagging performance as training material may not be sufficient to estimate them adequately.

The LSW tagger is an improved version of the sliding window (SW) PoS tagger \cite{Enrique-04}, and the main goal of the LSW tagger is to reduce the parameters of a SW tagger, by using approximations for the parameter estimation, without a significant loss in accuracy. Therefore, we briefly describe the SW tagger first, and then the LSW tagger.

\subsection{The SW tagger}
\subsubsection{Overview}

Let $\Gamma = \{ \gamma_{1}, \gamma_{2}, \dots, \gamma_{| \Gamma |} \}$ be the tag set, and $W = \{w_{1}, w_{2}, \dots\}$ be the words to be tagged. A partition of $W$ is established so that $w_{i} \equiv w_{j}$ if and only if both are assigned the same subset of tags, where each class of the partition is called an ambiguity class. Let $\Sigma = \{\sigma_{1}, \sigma_{2}, \dots, \sigma_{| \Sigma |}\}$ be the collection of ambiguity classes, where  each $\sigma_{i}$ is an ambiguity class. Let $T : \Sigma \rightarrow 2^{\Gamma}$ be the function returning the collection $T(\sigma)$ of PoS tags for an ambiguity class $\sigma$.

The PoS tagging problem may be formulated as follows: given a text $w[1]w[2] \dots w[L] \in W^{+}$, each word $w[t]$ is assigned (using a lexicon and a morphological analyzer) an ambiguity class $\sigma[t] \in \Sigma$ to obtain the ambiguously tagged text $\sigma[1] \sigma[2] \dots \sigma[t] \in \Sigma^{+}$; the task of a PoS tagger is to obtain a tag sequence $\gamma[1] \gamma[2] \dots \gamma[t] \in \Gamma^{+}$ as correct as possible, that is, the one that maximizes the probability of that tag sequence given the word sequence:
\begin{equation}
\gamma^{*}[1]\dots\gamma^{*}[L] = \argmax_{\gamma[t] \in T(\gamma[t])} P(\gamma[1]\dots\gamma[L] | \sigma[1]\dots\sigma[L])
\end{equation}
The core idea of SW PoS tagging is to use the ambiguity classes of neighboring words to approximate the dependencies locally:
\begin{equation}
P(\gamma[1]\dots\gamma[L] | \sigma[1] \dots \sigma[L]) = \prod_{t = 1}^{t = L} p(\gamma[t] | C_{(-)}\sigma[t]C_{(+)})
\end{equation}
where $t = 1 \dots L$, $C_{(-)}$ is the left context of length $N_{(-)}$ (e.g. if $N_{(-)} = 1$, then $C_{(-)} = \gamma[t-1])$, and $C_{(+)}$ is the left context of length $N_{(+)}$.

\subsubsection{Unsupervised parameter estimation}

Let $p(\gamma|C_{(-)} \sigma C_{(+)})$ be the probability of  a tag $\gamma$ appearing between the context $C_{(-)}$ and $C_{(+)}$. The most probable tag $\gamma^{*}[t]$ is selected as the one with the highest probability by the formula:
\begin{equation}
\gamma^{*}[t] = \argmax_{\gamma \in T(\sigma [t])} p(\gamma | C_{(-)} \sigma C_{+)})
\end{equation}
Estimating the parameters from a tagged corpus would be straightforward, but estimating from an untagged corpus requires an iterative process. Let ${\tilde{n}}_{C_{(-)} \gamma C_{(+)}}$ (a simpler and interchangeable representation for $p(\gamma|C_{(-)} \sigma C_{(+)})$ ) be the effective number of times (count) that $\gamma$ appears between the context $C_{(-)}$ and $C_{(+)}$. Following the steps in \cite{Enrique-04}, we can estimate ${\tilde{n}}_{C_{(-)} \gamma C_{(+)}}$ iteratively by:
\begin{equation}
\begin{split}
&{\tilde{n}}_{C_{(-)} \gamma C_{(+)}}^{[k]} = \\ &{\tilde{n}}_{C_{(-)} \gamma C_{(+)}}^{[k - 1]} \sum_{\sigma : \gamma \in T(\sigma)} n_{C_{(-)} \sigma C_{(+)}} {\left(\sum_{\gamma' \in T(\sigma)} {\tilde{n}}_{C_{(-)} \gamma' C_{(+)}}^{[k - 1]}\right)}^{-1}
\end{split}
\end{equation}
A recommended initial value could be obtained by assuming that all the tags $\gamma$ in $\sigma$ are equally probable.

\subsection{The LSW tagger}
\subsubsection{Overview}
The SW tagger tags a word by looking at the ambiguity classes of neighboring words, and has therefore a number of parameters in $O({|\Sigma|}^{N_{(-)} + N_{(+)}} |\Gamma|)$. The LSW tagger \cite{Enrique-05} tags a word by looking at the possible tags of neighboring words, and therefore it has a number of parameters in  $O({|\Gamma|}^{N_{(-)} + N_{(+)} + 1})$. Usually the tag set size $|\Gamma|$ is significantly smaller than the combinational ambiguity class size $|\Sigma|$. In this way, the number parameters is effectively reduced.

The LSW approximates the best tag as follows:
\begin{equation}
\begin{split}
&\gamma^* = \argmax_{\gamma \in T(\sigma[t])} \\ &\sum_{E_{(-)} \in T'(C_{(-)}[t]) \atop E_{(+)} \in T'(C_{(+)}[t])} p(E_{(-)} \gamma E_{(+)} | C_{(-)}[t] \gamma[t] C_{(+)}[t])
\end{split}
\end{equation}
where $T' :  \Sigma^* \rightarrow 2^{\Gamma^*}$, an extension of $T$, returns the set of tag sequences for an ambiguity sequence; $E_{(-)}$ and $E_{(+)}$ are the left and right tag sequence respectively.

\subsubsection{Unsupervised parameter estimation}
Following a procedure similar to that for the SW tagger, we can derive an iterative process to train the LSW tagger.
\begin{equation}
\begin{split}
&{\tilde{n}}_{E_{(-)} \gamma E_{(+)}}^{[k]} = {\tilde{n}}_{E_{(-)} \gamma E_{(+)}}^{[k - 1]} \sum_{\sigma : \gamma \in T(\sigma) \atop {C_{(-)}: E_{(-)} \in T'(C_{(-)}) \atop C_{(+)}: E_{(+)} \in T'(C_{(+)})}}  \\ &n_{E_{(-)} \sigma E_{(+)}} {\left(\sum_{\gamma' \in T(\sigma) \atop {C_{(-)}: E_{(-)} \in T'(C_{(-)}) \atop C_{(+)}: E_{(+)} \in T'(C_{(+)}) }} {\tilde{n}}_{E_{(-)} \gamma' E_{(+)}}^{[k - 1]}\right)}^{-1}
\end{split}
\end{equation}

where ${\tilde{n}}_{E_{(-)} \gamma E_{(+)}}$ is the effective number of times (count) that $\gamma$ appears between the context of tags $E_{(-)}$ and $E_{(+)}$.

Similarly to the initialization step in the SW tagger, a recommended initial value can be obtained by assuming that all the tag sequences $E_{(-)} \gamma E_{(+)}$ in the window $C_{(-)} \sigma C_{(+)}$ are equally probable.

\subsection{LSW with forbid and enforce rules}
There are forbid and enforce rules for sequences of two PoS tags in the current implementation of the Apertium PoS tagger. They were successfully applied in the original HMM tagger in Apertium, with a significant improvement in accuracy \cite{Sheikh-09}, simply by making the corresponding transition probabilities equal to zero. The SW tagger could not make use of forbid and enforce rules because of the fact that it works with ambiguity classes, while on the other hand, the LSW tagger can easily incorporate them as it works directly with PoS tags

The rules can be introduced right after the initialization step. For a tag sequence in the parameter space, if any consecutive two tags match a forbid rule or fail to match an enforce rule, the underlying parameter ${\tilde{n}}_{E_{(-)} \gamma E_{(+)}}$ will be given a starting value of zero.

In this way, for an LSW tagger with rules, the initial value could be given as follows,
\begin{equation}
{\tilde{n}}_{E_{(-)} \gamma E_{(+)}}^{[0]} = 
  \begin{cases}
    0 & \text{if $E_{(-)} \gamma E_{(+)}$ is not valid,} \\
    \Lambda & \text{otherwise}
  \end{cases}
\end{equation}
where
\begin{equation}
\Lambda = \sum_{\sigma : \gamma \in T(\sigma) \atop {C_{(-)}: E_{(-)} \in T'(C_{(-)}) \atop C_{(+)}: E_{(+)} \in T'(C_{(+)})}} n_{E_{(-)} \sigma E_{(+)}} \frac {1} {|V'(C_{(-)} \sigma C_{(+)})|}
\end{equation}
where, the validity of $E_{(-)} \gamma E_{(+)}$ is determined by forbid and enforce rules, and the function $V'$ returns the collection of valid (enforced or not forbidden) tag sequences contained in the ambiguity class sequence $C_{(-)} \sigma C_{(+)}$.

\section{Experiments}
\subsection{Training data and test set}

The experiments are conducted on three languages: Spanish (\texttt{apertium-en-es-0.8.0}), Catalan (\texttt{apertium-es-ca-1.1.0}), and English (\texttt{apertium-en-es-0.8.0}). We obtain the training data for Spanish and English by sampling text from the Europarl corpus \cite{koehn2005epc}, and for Catalan by sampling text from the Catalan Wikipedia. The statistics on the training data and test data are shown in Table \ref{tab1}. Test data for Catalan and Spanish come from \texttt{apertium-es-ca-1.1.0}. It is worth noting that the English test set has been built by mapping the results form the TnT \cite{Thorsten-00} tagger as an approximation.

\begin{table}
\begin{center}
\begin{small}
\begin{tabular}{|c|c|c|c|}
  \hline
  Items	& Spanish & Catalan & English \\
  \hline
  \hline
  Words (train)             & 3 million & 4 million & 3 million \\
  \hline
  Amb. classes (train)      & 106       & 92	    & 68        \\
  \hline
  Words (test)              & 25, 000   & 25, 000   & 30, 000   \\
  \hline
  Amb. rate (test)          & 22.81\%   & 31.13\%   & 29.97\%   \\
  \hline
  Forbid rules              & 545       & 272	    & 117       \\
  \hline
  Enforce rules             & 15        & 25	    & 41        \\
  \hline  
\end{tabular}
\end{small}
\caption{Major statistics for the training and test data.}
\label{tab1}
\end{center}
\end{table}

\subsection{The LSW tagger vs. the SW tagger}

\begin{figure}[h]
\begin{center}
\includegraphics[scale=0.5]{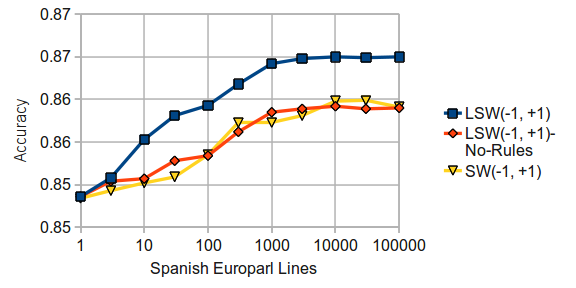}
\includegraphics[scale=0.5]{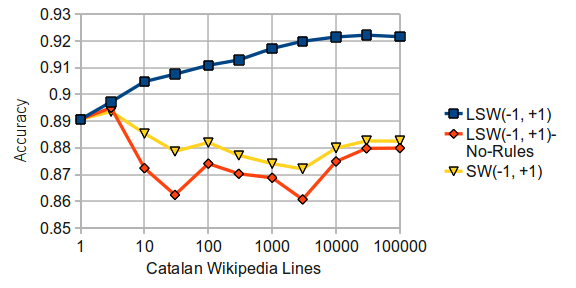}
\includegraphics[scale=0.5]{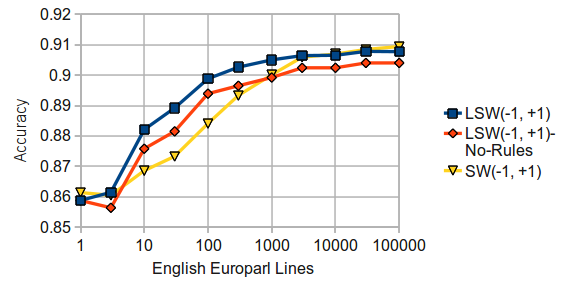}
\caption{Performance evaluation for (1) the LSW(-1, +1) tagger, (2) the LSW(-1, +1) tagger without rules, denoted as LSW(-1, +1)-No-Rules, and (3) the SW(-1, +1) tagger, all on Spanish, Catalan, and English. }
\label{fig1}
\end{center}
\end{figure}

We firstly study whether there is a difference between the LSW tagger and the SW tagger, keeping all other settings the same. Then we study whether rules can help improve the accuracy for the LSW tagger. ``Accuracy'' in the graph refers to the tagging precision of a tagger on the hand-tagged test set. Figure \ref{fig1} shows that rules help significantly for improving accuracy, and that the SW tagger behaves similarly to the LSW tagger without rules, which is consistent with the conclusion in \cite{Enrique-05}.

\subsection{Different window settings for the LSW tagger}

\begin{figure}[h]
\begin{center}
\includegraphics[scale=0.5]{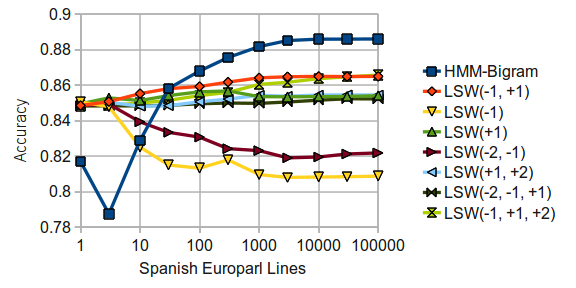}
\includegraphics[scale=0.5]{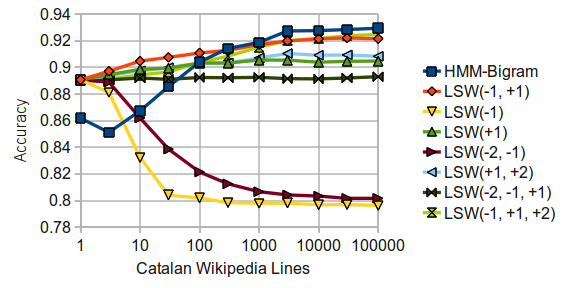}
\includegraphics[scale=0.5]{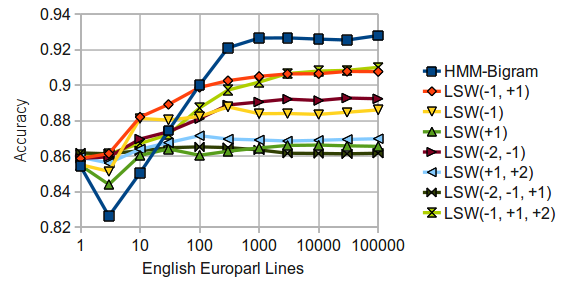}
\caption{Different window settings and their performance, tested on Spanish, Catalan, and English.}
\label{fig2}
\end{center}
\end{figure}

We study the performances of the LSW tagger with different window settings, and of the HMM tagger, on the three languages, as shown in Figure \ref{fig2}. We can see that the HMM tagger performs best among all the taggers, especially when there is enough training data. However, when training data is limited, the LSW taggers learn faster (need less words to learn) and more stably than the HMM tagger.

Among all the LSW taggers, the LSW(-1, +1), i.e. left context 1 and right context 1, performs best. When there are enough training data, the performances of the HMM tagger and the LSW(-1, +1) tagger are quite close.

Note that under some window settings, the performances of the LSW taggers even decrease as more training lines were added, e.g. LSW(-1) and LSW(-2, -1) for Spanish and Catalan. This is an unexpected phenomenon, and the reason for it would require further investigation.

\subsection{Using Constraint Grammar rules to support the HMM and LSW}

\begin{figure}[h]
\begin{center}
\includegraphics[scale=0.5]{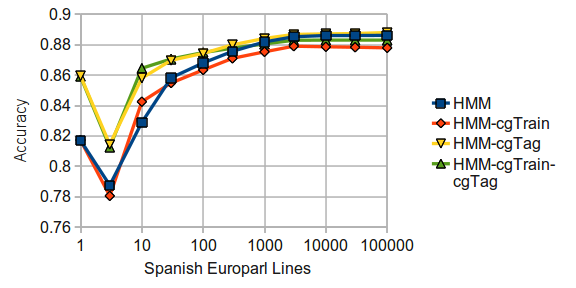}
\includegraphics[scale=0.5]{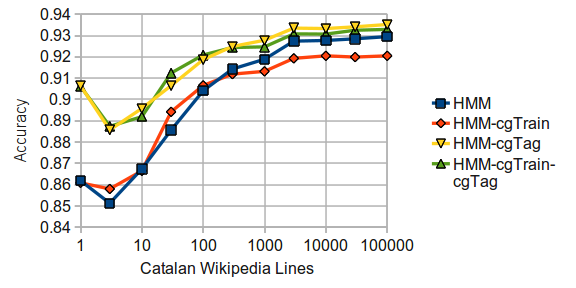}
\includegraphics[scale=0.5]{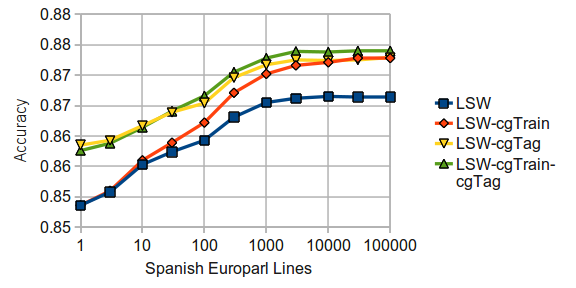}
\includegraphics[scale=0.5]{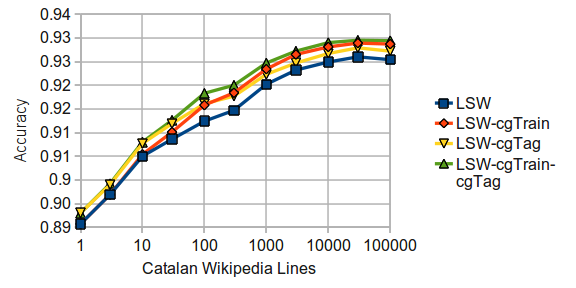}
\caption{Performance evaluation for HMM and LSW with and without CG.}
\label{fig3}
\end{center}
\end{figure}

We also tested whether the use of Constraint Grammar (CG) rules helps to improve the accuracy obtained by both HMM and LSW taggers, along the lines suggested in \cite{conf/lrec/HuldenF12}. For that, we used the CG rules already present in Apertium packages \texttt{apertium-eo-es-0.8.2} for Spanish and \texttt{apertium-eo-ca-0.8.2} for Catalan respectively (a CG module is integrated in many Apertium language pairs). Figure \ref{fig3} shows that CG helps almost in all settings. It is also shown that CG rules help the two taggers in different situations: for the HMM tagger, the positive contribution of CG rules is larger when training data is limited than when training data is relatively enough; while for the LSW tagger, the trend is almost the opposite, that CG rules contribute even more when training data is relatively enough. Note that the logical approach would be to use CG rules both for reducing ambiguity for the training corpus (denoted as \textbf{cgTrain} in Figure \ref{fig3}) and for reducing ambiguity right after morphological analyzer and before the PoS tagger (denoted as \textbf{cgTag} in Figure \ref{fig3}); the results are however almost indistinguishable from those obtained applying CG in either step.

\section{Discussion and future work}

We reviewed the mechanism and unsupervised parameter estimation methods for both the SW and LSW taggers. Compared with previous work \cite{Enrique-04,Enrique-05}, firstly, we proposed a method for incorporating the forbid and enforce rules already used for HMM taggers in Apertium into the LSW tagger; and secondly, the implementation is the first time that the LSW tagger is integrated into a real machine translation system (Apertium), and at the same time, its code is free/open-source.

We also conducted experiments to compare the performances of the LSW tagger  with different settings, and with respect to the original HMM tagger. Firstly, the HMM tagger performs slightly better than the LSW(-1, +1) tagger when there is enough training data, while the LSW(-1, +1) tagger learns faster and is more stable when training data is limited. Secondly, the LSW(-1, +1) tagger performs best among all the other window settings, and better than the SW(-1, +1) tagger, which behaves similarly with LSW(-1, +1)-No-Rules. Thirdly, we have found that the use of CG rule sets already existing in some Apertium taggers helps significantly to improve accuracy based both on the HMM and LSW taggers, and that for the HMM tagger CG rules help more when training data is limited, while for the LSW tagger CG rules help more when training data is relatively enough.

The reason why the performance of the LSW tagger under some window settings worsens as more training lines are added also requires more efforts to study. Source code is available through the Apertium Subversion repository\footnote{\url{https://svn.code.sf.net/p/apertium/svn/branches/apertium-swpost}} under a free/open-source license.

\paragraph{Acknowledgements:}
Support from Google Summer of Code (summer scholarship for Gang Chen) and from the Spanish Ministry of Economy and Competitiveness through grant TIN2012-32615 are gratefully acknowledged. The authors also thank Francis M.\ Tyers and Jim O'Regan for useful comments.

\bibliographystyle{lrec2006}
\bibliography{LSW}

\end{document}